%% file: main.tex
\documentclass[sigconf,natbib=false]{acmart}
\AtBeginDocument{%
  \providecommand\BibTeX{{%
    \normalfont B\kern-0.5em{\scshape i\kern-0.25em b}\kern-0.8em\TeX}}}

\setcopyright{none}
\renewcommand\footnotetextcopyrightpermission[1]{}
\input{text/acrons}

\begin{document}

\title{The Role of Machine Learning for Trajectory Prediction in Cooperative Driving}

\author{Luis Sequeira}
\email{luis.sequeira@kcl.ac.uk}
\affiliation{%
  \institution{Centre For Telecommunications Research \\Department of Engineering}
%   \streetaddress{King's College London}
  \city{King's College London}
}

\author{Toktam Mahmoodi}
\email{toktam.mahmoodi@kcl.ac.uk}
\affiliation{%
  \institution{Centre For Telecommunications Research \\Department of Engineering}
%   \streetaddress{King's College London}
  \city{King's College London}
}

% \renewcommand{\shortauthors}{Sequeira and Mahmoodi}

\begin{abstract}
\input{text/abstract}
\end{abstract}

%   <concept>
%       <concept_id>10010147.10010257.10010258.10010261</concept_id>
%       <concept_desc>Computing methodologies~Reinforcement learning</concept_desc>
%       <concept_significance>500</concept_significance>
%       </concept>
%   <concept>
%       <concept_id>10003033.10003099.10003100</concept_id>
%       <concept_desc>Networks~Cloud computing</concept_desc>
%       <concept_significance>300</concept_significance>
%       </concept>
%   <concept>
%       <concept_id>10010147.10010257</concept_id>
%       <concept_desc>Computing methodologies~Machine learning</concept_desc>
%       <concept_significance>500</concept_significance>
%       </concept>
%  </ccs2012>

% \ccsdesc[500]{Computing methodologies~Reinforcement learning}
% \ccsdesc[300]{Networks~Cloud computing}
% \ccsdesc[500]{Computing methodologies~Machine learning}



\maketitle

\section{Introduction}
\label{Introduction}
\input{text/intro}

\section{State of the Art}
\label{State of the Art}
\input{text/soa}

\section{Cooperative Lane Merge Model} 
\label{Model}
\input{text/model}

\section{Results}
\label{Results}
\input{text/results}

\section{Conclusion}
\input{text/conclusion}
\label{Conclusion}

\begin{acks}
\input{text/acks}
\end{acks}

% \bibliographystyle{ACM-Reference-Format}
% \bibliographystyle{biblatex}
% \citestyle{acmnumeric}
% \setcitestyle{numbers,sort&compress}
% \bibliography{text/references}
\printbibliography

\end{document}

%% file: text/acrons.tex
\usepackage[backend=biber,
style=numeric,
sorting=none]{biblatex}
\usepackage{multicol}
\usepackage{multirow}
\usepackage{colortbl}
\usepackage{tabularx}
\usepackage{siunitx}
\usepackage{lscape}
\usepackage{pdflscape}
\usepackage{booktabs}
\usepackage{lipsum}
\usepackage{gensymb}
\usepackage{amsmath}
\usepackage[]{algorithm2e}
\usepackage{float}
\usepackage[acronym,nomain,nonumberlist]{glossaries}
\usepackage{subfig}
\definecolor{row1}{rgb}{0.95,0.95,0.95}
\definecolor{row2}{rgb}{0.85,0.85,0.85}
\newacronym{5g}{$5G$}{$5^{th}$ Generation Mobile Network} 
\newacronym{cacc}{CACC}{Cooperative Adaptive Cruise Control}
\newacronym{etsi}{ETSI}{European Telecommunications Standards Institute}
\newacronym{it}{IT}{Information Technologies}
\newacronym{its}{ITS}{Intelligent Transport System} 
\newacronym{iot}{IoT}{Internet of Things}
\newacronym{mec}{MEC}{Multi-Access Edge Computing}
\newacronym{ran}{RAN}{Radio Access Network}
\newacronym{sdn}{SDN}{Software Defined Networking}
\newacronym{v2x}{V2X Gateway}{Vehicle-to-Everything}
\newacronym{v2v}{V2V}{Vehicle-to-Vehicle}
\newacronym{v2e}{V2X}{Vehicle-to-Everything}
\newacronym{5gaa}{$5$GAA}{$5G$ Automotive Association}
\newacronym{json}{JSON}{JavaScript Object Notation}
\newacronym{tcp}{TCP}{Transmission Control Protocol}
\newacronym{udp}{UDP}{User Datagram Protocol}
\newacronym{rud}{RUD}{Road User Description}
\newacronym{id}{ID}{Identification}
\newacronym{rmse}{RMSE}{Root-Mean-Square Error}
\newacronym{gui}{GUI}{Graphical User Interface}
\newacronym{gdm}{GDM}{Global Dynamic Map}
\newacronym{uuid}{UUID}{Universally Unique Identifier}
\newacronym{lstm}{LSTM}{Long Short-Term Memeory}
\newacronym{rl}{RL}{Reinforcement Learning}
\newacronym{rfc}{RFC}{Random Forest Classifier}
\newacronym{dqn}{DQN}{Deep Q-Network}
\newacronym{ngsim}{NGSIM}{Next Generation Simulation Model}
\newacronym{to}{TO}{Traffic Orchestrator}
\newacronym{ddqn}{Double-DQN}{Double Deep Q-Network}
\newacronym{dueling-dqn}{Dueling DQN}{Dueling Deep Q-Network}
\newacronym{sarsa}{SARSA}{State-Action-Reward-State-Action}
\newacronym{kpi}{KPI}{Key Performance Indicator}
\newacronym{gelf}{GELF}{Graylog Extended Log Format}
\newacronym{df}{DF}{Data Fusion}
\newacronym{ecdf}{ECDF}{Empirical Cumulative Distribution Function}
\newacronym{rtt}{RTT}{Round Trip Time}
\newacronym{ai}{AI}{Artificial Intelligence}
\newacronym{ml}{ML}{Machine Learning}

%% file: text/abstract.tex
In this paper, we study the role that machine learning can play in cooperative driving. Given the increasing rate of connectivity in modern vehicles, and road infrastructure, cooperative driving is a promising first step in automated driving. The example scenario we explored in this paper, is coordinated lane merge, with data collection, test and evaluation all conducted in an automotive test track. The assumption is that vehicles are a mix of those equipped with communication units on board, i.e. connected vehicles, and those that are not connected. However, roadside cameras are connected and can capture all vehicles including those without connectivity. We develop a \textit{Traffic Orchestrator} that suggests trajectories based on these two sources of information, i.e. connected vehicles, and connected roadside cameras. Recommended trajectories are built, which are then communicated back to the connected vehicles. We explore the use of different machine learning techniques in accurately and timely prediction of trajectories.

%% file: text/intro.tex
With the advances of mobile communication in \gls{5g} to serve beyond broadband users, and address the needs of more critical infrastructure from industrial networks \cite{virtuwind-intro} to automotive \cite{5gcar-intro}, new connectivity era came to live. Additional features of \gls{5g} radio in short range and high rate communication with mmWave \cite{v2x-mmwave}, and ability of direct communication between vehicles (Vehicle-to-Vehicle, or V2V) as addressed under 5G Vehicle-to-Everything (V2X) communications, enabled the long vision of fully connected transportation. Connected vehicles, on the other hand, provide a rich data platform in \gls{its}. Such data has resulted in enhancements to road safety, traffic efficiency, improvement in environmental impacts and energy costs \cite{its4}. A connected vehicle, or in other words, a vehicle capable of transmitting and receiving data to and from the network can potentially increase awareness of the driver (or the driving agent). 

In this paper, we explore the role of different \gls{ml} techniques in order to predict trajectories in a cooperative lane merge scenario. The edge cloud deployment is based on a central entity that collects data from vehicles and roadside cameras, and performs the prediction. We evaluate two different approaches, the first one based on \gls{ml} classifiers where four algorithms are considered: Random Forest, K-Nearest Neighbours, Decision Tree and Gradient Boosting. The second approach is based on using \gls{rl}. The main goal is to predict the most appropriate manoeuvre for all the vehicles involved, so that the merging vehicle can execute the manoeuvre safely. 

The work from \cite{lm13} was used as a starting point. In this paper, the architecture for the \textit{Traffic Orchestrator} was enhanced with three main changes: $1)$ a new message format to communicate with the latest versions of external components, $2)$ a mechanism for logging and monitoring and $3)$ an alignment with a service-oriented architecture. On the other hand, a new \gls{dueling-dqn} model was implemented and evaluated. Furthermore, real-world tests were conducted in a test track\footnote{Tests and evaluations were conducted in UTAC CEARM (Paris) site: https://www.utacceram.com/proving-grounds} and a performance evaluation of a cooperative lane merge scenario is presented. The remainder of this paper is organised as follows. Section \ref{State of the Art} provides the state of the art for lane merge algorithms. Section \ref{Model} presents the system model for cooperative lane merge. Section \ref{Results} presents the data preparation methods, and the performance evaluation of different \gls{ml} models. Finally, the paper is concluded in Section \ref{Conclusion}.

%% file: text/soa.tex
In a collision free merge, a certain safety distance between the merging vehicle and the other vehicles is required. A problem arises when the gap between vehicles on the merging lane is not enough for the merging vehicle to fit in between. Therefore, a lane merge coordination algorithm is needed to perform actions on merging vehicles providing successful and safe lane merges \cite{lm8}. 

In the last few years, significant progress has been made in solving this challenging problem. In \cite{lm9}, a representation of the on-road environment (Dynamic Probabilistic Drivability Map) was presented which delivers cost effective recommendations based on dynamic programming. The theoretical formulation of this work was tested with data from $40$ real-world merges. In \cite{lm7}, a work-in-progress using Long Short-Term Memory architecture with Deep Q-Learning was presented. The scenario considers an on-ramp merging involving three vehicles: the merging vehicle and two vehicles on the mainline. The work considers as input variables: speed, position, heading angle, and distances to the right and left lane. The algorithm has not been verified or validated.

The work in \cite{autonomousDrivingRl} adopts a deep \textit{\gls{rl}} approach handling input values from camera and laser sensor the vehicle owns. A monolith architecture embedded in the vehicle was proposed, so that every vehicle that includes the equipment specified would have to generate the calculations. In \cite{bayesForLaneChange}, the authors used Bayes Classifiers and Decision Trees to predict if a vehicle can merge. Although the results proved accurate in some cases, the approach did not detail the lane merge trajectory that should be used. 

A discussion of the assessment that needs to be carried out before Deep Learning can be considered in autonomous vehicles is presented in \cite{deepRlforVehicles}. \textit{\gls{rl}} only guarantees a convergence for an optimal value function if every state is visited infinite number of times. For starters, a large amount of data would be needed in order to simulate visiting the states an infinite number of times, obviously infeasible due to time constrains. However, with some work done in function approximation, the reward can be calculated with the state and action only, thus minimising the storage needed to hold the infinite sets of combinations, which is where the deep learning component alleviates the problem by generalising the approximation function. Reward function also provides a way for the agent to prioritise tasks codifying necessary behaviour to an agent, which is vital for an optimal function to be learned \cite{rewardFunctionDesc}.

%% file: text/model.tex
\begin{figure}[!t]
    \centering
    \includegraphics[width=\columnwidth]{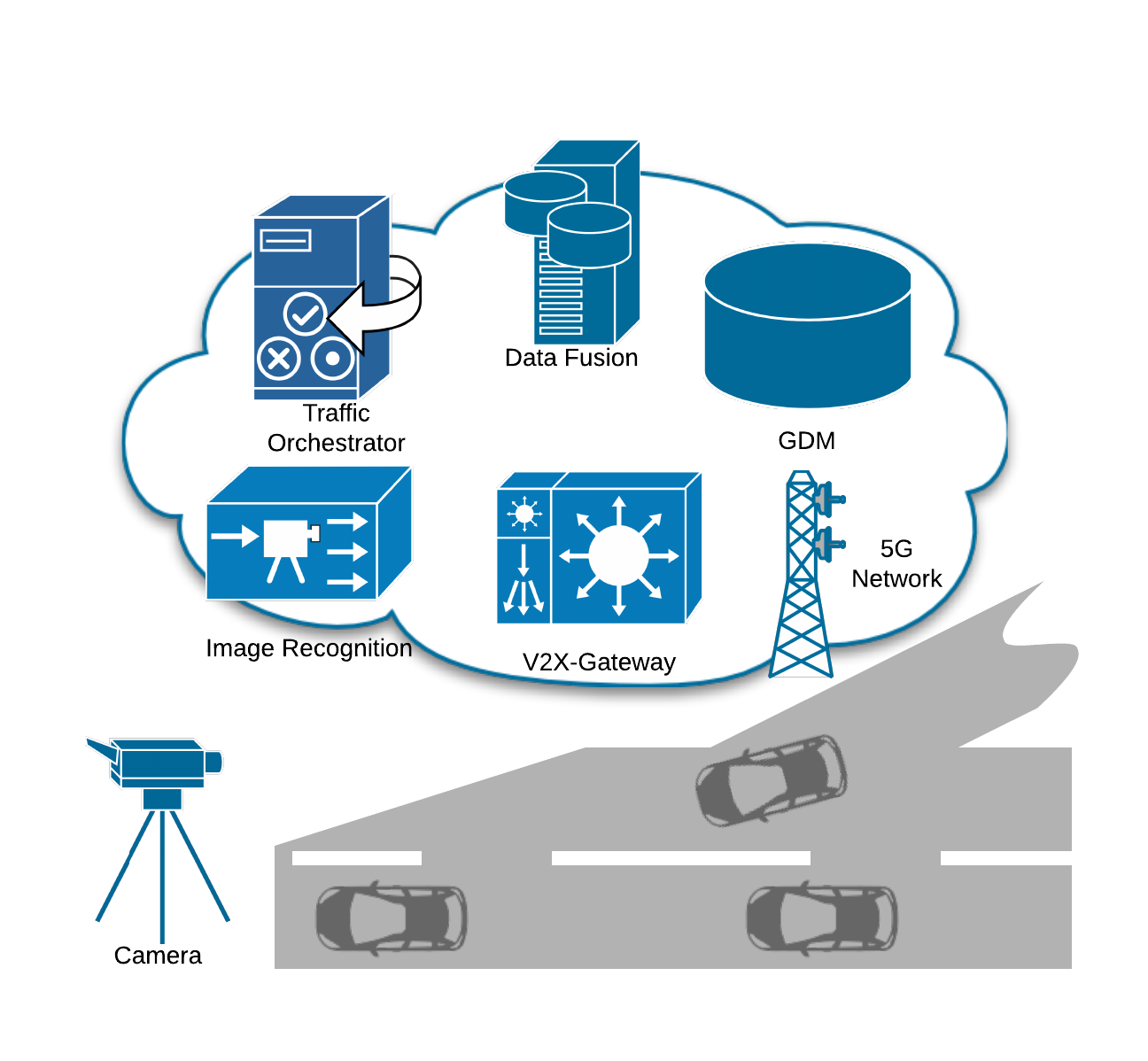}
    \caption{Lane merge coordination scenario.}
    \label{fig:LaneMergeCoordination}
\end{figure}

The cooperative lane merge scenario examined in this work is depicted in Fig. $\ref{fig:LaneMergeCoordination}$, where a connected vehicle will attempt to merge onto a carriageway in which connected and unconnected vehicles are also present. The general architecture aggregates data for the cameras (regarding connected and unconnected vehicles) and from the connected vehicles themselves, with such as data, the coordination model is able to react to road changes and to predict trajectory recommendations. Through an edge cloud approach, bespoke trajectory recommendations are sent by central coordination mechanism to connected vehicles. The model consists of $5$ different components: a \textit{\gls{v2x}}, an \textit{Image recognition} system, a \textit{\gls{gdm}}, a \textit{Data Fusion} and a \textit{Traffic Orchestrator}.

The \textit{\gls{v2x}} is a context-based messaging system who is responsible for forwarding messages among components in the architecture. The \textit{\gls{v2x}} also provides communication with connected vehicles by means of the \textit{\gls{5g} Network}. The \gls{5g} communication seeks to maintain the physical connection between the \textit{\gls{v2x}} and the connected vehicles, but also allows edge computing, network slicing, and quality of service. Furthermore, the \textit{\gls{v2x}} enforces a subscription based model for applications, such as the \textit{Traffic Orchestrator}, to receive messages regarding vehicular description and trajectory information. 

The main goal of the \textit{Image recognition} system \cite{kai} is to deliver data regarding unconnected vehicles, however, it collects information about all the vehicles on the road (connected and unconnected ones) in a specified area. In this sense, the data delivered by the \textit{Image recognition} system is partially duplicated, since connected vehicles share their own data too. The data shared by connected vehicles and the \textit{Image recognition} system includes the localisation and trajectory-based parameters. The \textit{Image recognition} sends the obtained data to the \gls{v2x}, which in turn forwards the messages to the \textit{\gls{gdm}}. The \textit{\gls{gdm}} stores environmental information about connected and unconnected vehicles in a database. The \textit{\gls{gdm}} ensures that stored data is up to date. The \textit{Data Fusion} provides a synchronisation mechanism for data originating from different sources (e.g., one from the \textit{Image recognition} system and another from a connected vehicle in a closely localised time frame, respectively). The \textit{Data Fusion} sends the information to applications that are subscribed to a specific location boundary. Additionally, it includes the monitoring and evaluation platform to assess communication \glspl{kpi}.

The \textit{Traffic Orchestrator} will process data regarding connected and unconnected vehicles to give rise to trajectories for connected vehicles. The \textit{Traffic Orchestrator} considers time-critical variables such as timestamp, vehicle location, speed and vehicle-specific dimensions. Once the \textit{Traffic Orchestrator} provides a coordinated trajectory recommendation for a single or set of road users, this trajectory will then be sent to the \textit{\gls{v2x}} forwarding them to the connected vehicles. The connected vehicles have the choice to either accept, reject or abort the recommendation. This feedback information can be used to recalculate trajectory recommendations. To this end, a set of messages was defined for communicating all the components within the lane merge coordination. Messages used in the communication will employ a common message formatting based on \gls{json}. This allows to communicate human-readable text, that can be received and processed in any software component. 

\begin{figure}[!t]
	\centering
	\includegraphics[width=0.9\columnwidth,height=9.5cm]{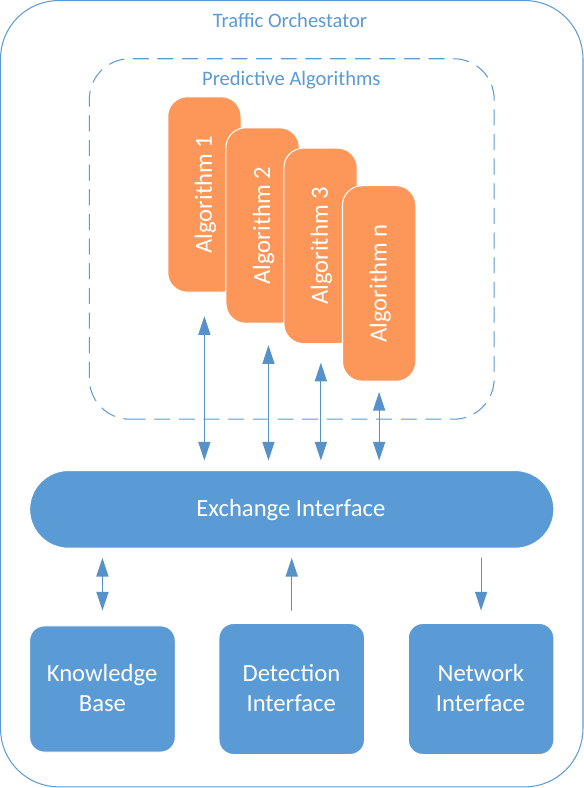}
	\caption{Proposed \textit{Traffic Orchestrator} architecture.}
	\label{fig:proposed_architecture}
\end{figure}

The proposed architecture for the \textit{Traffic Orchestrator} system is presented in Fig. \ref{fig:proposed_architecture}. The main purpose of the \textit{Detection Interface} is to receive data coming from the \textit{V2X-Gateway}, over a \gls{tcp} connection. The \textit{Detection Interface} reads \gls{json} strings and commences the process of converting the \gls{json} readable messages into more compact and computer efficient entities to feed the models. Similarly, the \textit{Network Interface} translates information within the \textit{Traffic Orchestrator}, to information that is readable and accepted by the \textit{V2X-Gateway}. A \textit{Knowledge Base} has been designed to store the information sent to the \textit{Traffic Orchestrator}. The \textit{Knowledge Base}, although simpler than a database, will have to mimic the access and modification functions of a typical database. It is maintained to guarantee that a manoeuvre recommendation is calculated based on all current road-environment knowledge. The \textit{Knowledge Base} will contain only the data that the \textit{\gls{gdm}} has most recently transmitted. This will prevent maintaining information within the \textit{Traffic Orchestrator} that is out-of-date or no longer relevant. The \textit{Exchange Interface} has two responsibilities: mediate the flow of information across all interfaces in the \textit{Traffic Orchestrator} and provide access for consistent methods to a set of \textit{Traffic Orchestrator} functionalities, allowing different algorithms to run on top of it.

%% file: text/results.tex
This section draws the experiences of using \gls{ml} for predicting trajectories in a cooperative driving environment. Initially, the data preparation process is highlighted, then two different approaches are presented. The first one is focused on the performance of different \gls{ml} classifiers including Random Forest, K-Nearest Neighbours, Decision Tree and Gradient Boosting \cite{lm13}. The second approach uses \gls{dueling-dqn}, and its implementation is further discussed here. Moreover, evaluations are performed in the test track from where we analyse the predicted trajectories based on the vehicles' positioning information (longitude/latitude) and their acceleration. Finally, we provide some lessons learnt regarding the design implementation and the overall demonstration. 

\subsection{Data Preparation}

\begin{figure*}[!t]
    \centering
    % \vspace{0.02in}
    \begin{tabular}{c}
    \subfloat[\textit{true} recommendation.]{
        \includegraphics[width=0.3\textwidth]{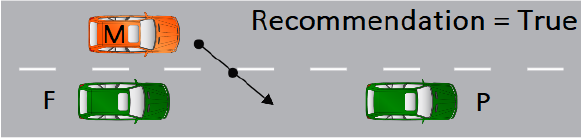}
        \label{fig:positiveRecommendation}
	}    
    \subfloat[\textit{false} recommendation.]{
        \includegraphics[width=0.3\textwidth]{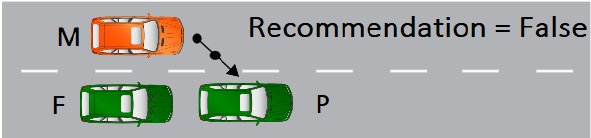}
        \label{fig:negativeRecommendation}
	} 
    \subfloat[Merging vehicle is ``behind" following vehicle.]{
        \includegraphics[width=0.3\textwidth]{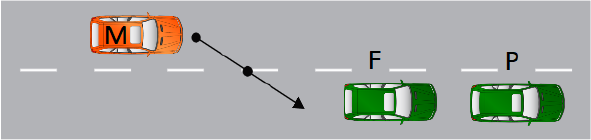}
        \label{fig:behindFollowing}
	}
    \end{tabular}
    \caption{Design premises for different road situations.}
    \label{fig:road_situation}
\end{figure*}

Two distinct datasets collected by Federal Highway Administration Research and Technology\footnote{\url{https://www.fhwa.dot.gov}} are adopted in this work. Theses datasets represent the data collected from two American motorways: $I-80$ and $US-101$. The dataset contains more than $10.8$ million rows with $25$ values per sample (e.g., vehicle identifier, coordinates, speed, acceleration, size and heading of the vehicle). 

We defined a lane merge instance using $3$ vehicles as shown in Fig. \ref{fig:road_situation}, whereby, the merging vehicle (denoted \textit{M}) has the goal of joining a new lane in between a preceding (denoted \textit{P}) and following (denoted \textit{F}) vehicle. This allows a more compact implementation which is conservative on terms of computing resources and model convergence time. Moreover, this abstraction permits to extrapolate to a more complex lane merge scenarios with different number of lanes on the road. We extracted every lane merge instance within the dataset along with their corresponding information. To obtain as much data as possible about a potential lane merge, data from $4$ seconds before and $3$ seconds after the lane change is detected and subsequently stored. Since the sample rate of the dataset is $10$ sample per second, a total of $70$ samples per lane merge instance were extracted. 

The extracted data was labelled as follow: if a lane merge is possible, the recommendation is \textit{true} (Fig. \ref{fig:positiveRecommendation}), otherwise \textit{false} (Fig. \ref{fig:negativeRecommendation}). All the cases in which the merging vehicle (M) is behind the following vehicle (F) on the new lane, are labelled as \textit{false} (Fig. \ref{fig:behindFollowing}), since they are not relevant. A safe distance between the front of the merging vehicle (M) and the back of the preceding vehicle (P) was maintained between the vehicles. Furthermore, a safe distance between the back of the merging vehicle (M) and the front of the following vehicle (F) was maintained between the vehicles. 

\subsection{First approach: \gls{ml} Classifiers}

\begin{table}[!t]
    \caption{Scores of different machine learning algorithms.}
    \label{table:ResultsML}
    \centering
    \scalebox{1}[1]{
    \begin{tabular}{lccc}
    	 \toprule
         \toprule
         Model & Merge & Acceleration & Heading \\
         \midrule
         Random Forest & $90.87$ & $75.74$ & $61.20$ \\
         K-Nearest Neighbours & $87.05$ & $-$ & $-$ \\
         Decision Tree & $86.84$ & $-$ & $-$ \\
         Gradient Boosting & $-$ & $76.55$ & $62.85$ \\
         \bottomrule
         \bottomrule
    \end{tabular} }
\end{table}

In order to evaluate the most suitable \gls{ml} classifier for predicting lane merges, acceleration and heading, $4$ different \gls{ml} classifiers are analysed, i.e., Random Forest, K-Nearest Neighbours, Decision Tree and Gradient Boosting. 

To detect a lane merge, we use previously prepared data which contain a list of measurements for each vehicle. The process stores the information about the initial lane number and compares that value with the current lane. A trajectory recommendation checker was implemented to guarantee safety. For the checker to determine whether a trajectory recommendation could be \textit{true}, it checks if the gap is wide enough to accommodate the merging vehicle. The recommended acceleration is calculated with three rules: 1) \textit{false} lane marge recommendations use the average speed to the first point at which a recommendation is \textit{true}, 2) any position before the merging point uses an average of the accelerations from the considered position, to the merging point and 3) for any position after the merging point, an average of the accelerations from the merging point to the considered location, is calculated. The heading for \textit{true} trajectory recommendations should lead the vehicle to the merging point. For \textit{false} recommendations, the vehicle should be led to the first location at which trajectory recommendations becomes \textit{true}.

To avoid over-fitting, two hyperparameters were adjusted: \textit{maximum\_depth} and \textit{number\_of\_estimators}. The values considered for the \textit{number\_of\_estimators} were: $1$, $2$, $5$, $10$, $20$, $35$, $50$, $75$ and $100$. The best value for the \textit{number\_of\_estimators} is $100$ for the $4$ algorithms. On the other hand, \textit{maximum\_depth} is individually adjusted for every algorithm. To select the \textit{maximum\_depth} when predicting lane merges, $30$ consecutive depths (ranging from $1$ to $30$) were considered for Random Forest and Decision Tree, and $50$ for K-Nearest Neighbours (ranging from $1$ to $50$). To select a proper \textit{maximum\_depth} for the Random Forest, we estimated the error on the validation and training sets for every maximum depth value. The value was chosen by comparing the results obtained on the validation set with the specified number for depth. The value of $16$ was the last for which the accuracy of the validation set was not worse than $1.5\%$ than the accuracy on the training set. The accuracy of the test set was the same as the accuracy on the validation set. The same technique was used to train the no over-fitting Decision Tree. The best value for which the model was not over-fitting was equal to $11$ and showed less than $1\%$ of difference with the validation set. The K-Nearest Neighbours algorithm has a different property: the lower the number of $K$, the higher the probability of over-fitting. To minimise the chance of over-fitting, $K$ was set to $50$. 

Table \ref{table:ResultsML} shows a summary of the obtained scores for predicting lane merges. Random Forest obtained the higher score among the classifiers when predicting merges. On the other hand, acceleration and heading did not achieve optimal values to be considered as options for testing with real vehicles on the test track. The scores were calculated using two functions (sklearn.metrics): \textit{accuracy\_score} and \textit{score}. The \textit{accuracy\_score} function uses multi-label classification and was used to estimate the scores on the validation dataset. The \textit{score} function returns the mean accuracy on the given test data and labels. This function was used to calculate scores on the training dataset. Moreover, for Random Forest \textit{cross\_val\_score} function (sklearn.model\_selection) was used. This function estimates a score using cross-validation techniques. 

\subsection{Second approach: \gls{dueling-dqn}}

Accuracy in acceleration and heading is essential for a safe lane merge manoeuvre, since an error on those predictions can lead vehicles towards an unwanted location, provoke accidents, produce unpleasant and risky manoeuvres by accelerating very quickly. Due to the low accuracy when predicting acceleration and heading by the \gls{ml} classifiers, we decided to focus on a second approach using \gls{rl}. So, \gls{dueling-dqn} was trained using the same dataset that was split into $3$ subsets: training, testing and validation where each of them with $70\%$, $20\%$ and $10\%$ of the size of the original dataset respectively. We used the validation subset to adjust the model and the test subset to check its performance. 

\begin{figure}[!t]
    \centering
    \subfloat[Negative Rewards \label{fig:dueling_rewards_a}]{\includegraphics[width=0.5\textwidth]{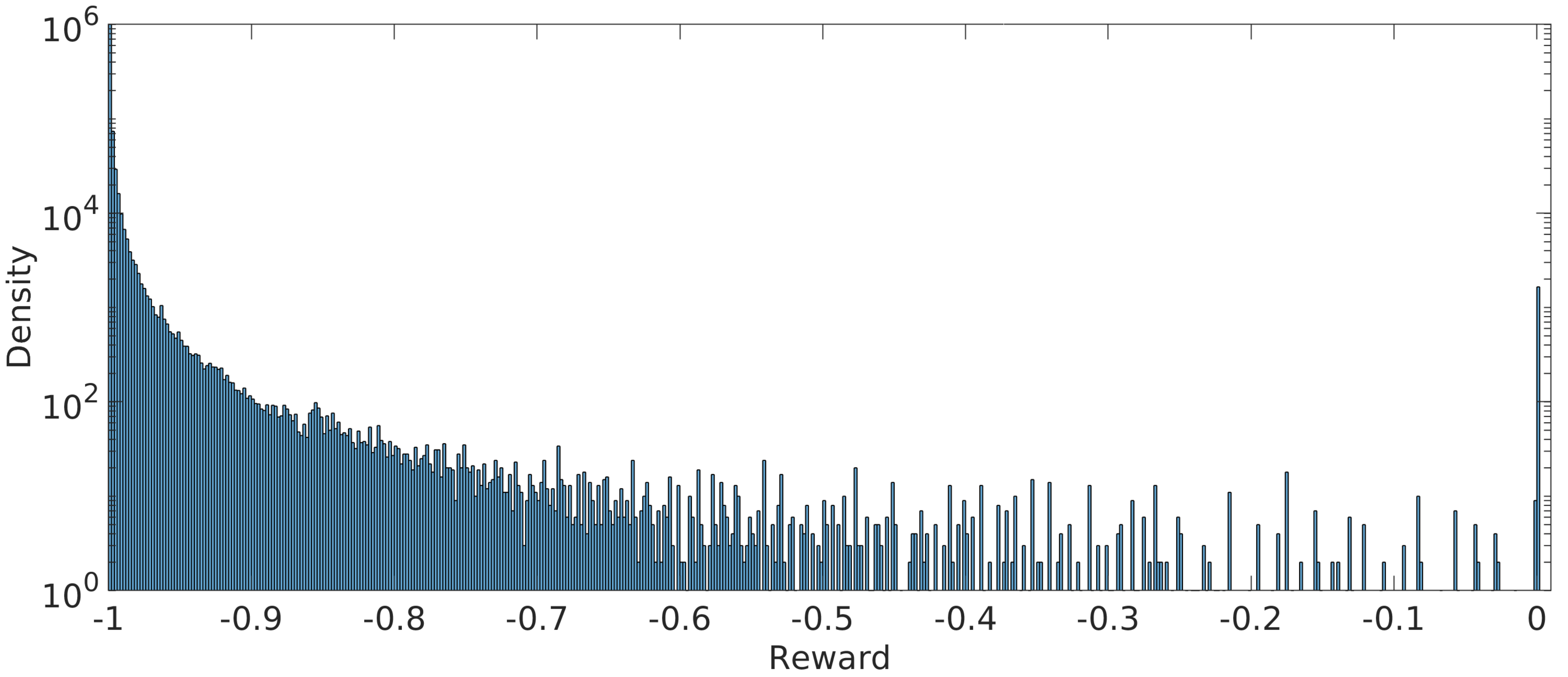}}
    \newline
    \subfloat[Positive Rewards \label{fig:dueling_rewards_b} ]{\includegraphics[width=0.5\textwidth]{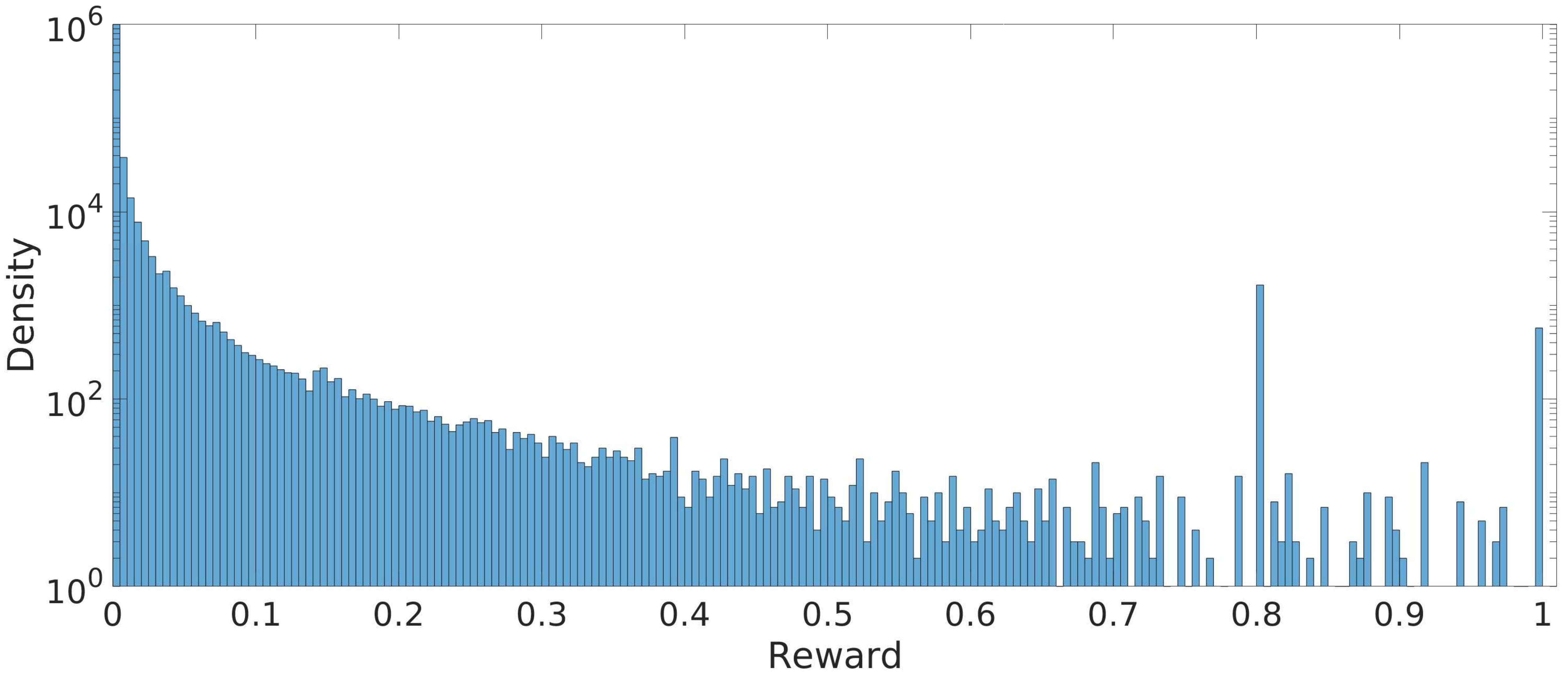}}
    \caption{Histogram for comparing assigned rewards for trajectory recommendation by \gls{dueling-dqn} agent.}
    \label{fig:comp_rew_ddqn}
\end{figure}

The reward function is calculated based on the vehicle position relative to the optimal merging point. A desired final state, is when the merging vehicle is successfully placed in between the following and preceding vehicle, while keeping a safety distance. To select the most appropriate reward function, two versions are evaluated: positive and negative reward. Fig. \ref{fig:comp_rew_ddqn} shows a histogram of positive and negative rewards assigned for each way-point in a trajectory recommendation during training time of the model. The positive and negative reward follow the same general pattern, but the positive reward has an increase in magnitude at a reward of $0.8$. The agent obtained a large density of rewards allocated at $0.8$ that proves the existence of a global minima in the positive reward function. The model needs to surpass this value in order to obtain a successful merge, which the negative reward function did not face. Based on that, the \gls{dueling-dqn} model with positive reward is more reliable.

\subsection{Cooperative Driving Evaluation in the Test Track}

\begin{figure}[!t]
    \centering
    \subfloat[UTAC test track for lane merge scenario \label{fig:real_scenario_b}]{\includegraphics[height=4cm,width=0.5\textwidth]{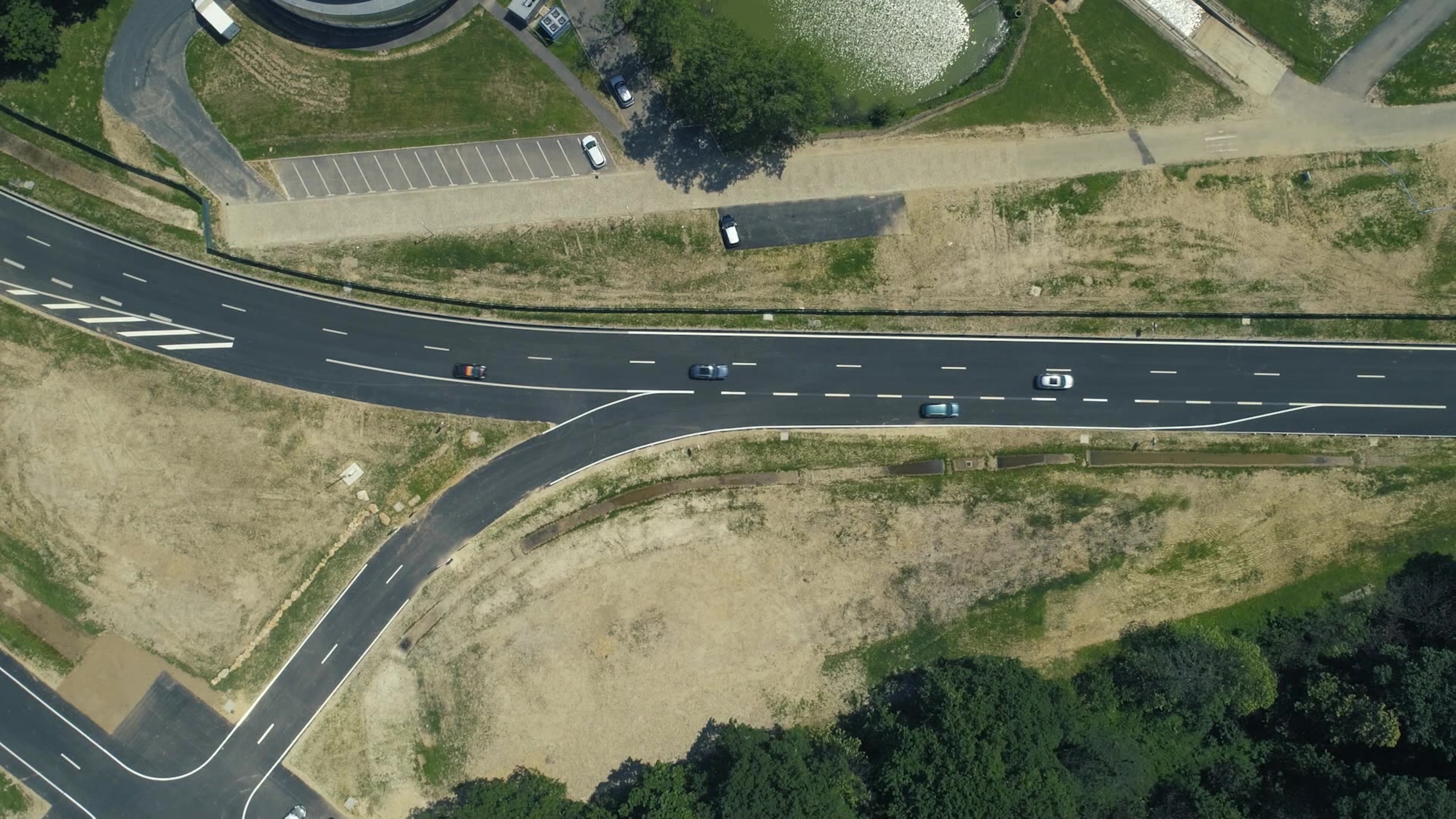}}
    \newline
    \subfloat[Human vs computed trajectories \label{fig:human_vs_to_real}]{\includegraphics[width=0.5\textwidth]{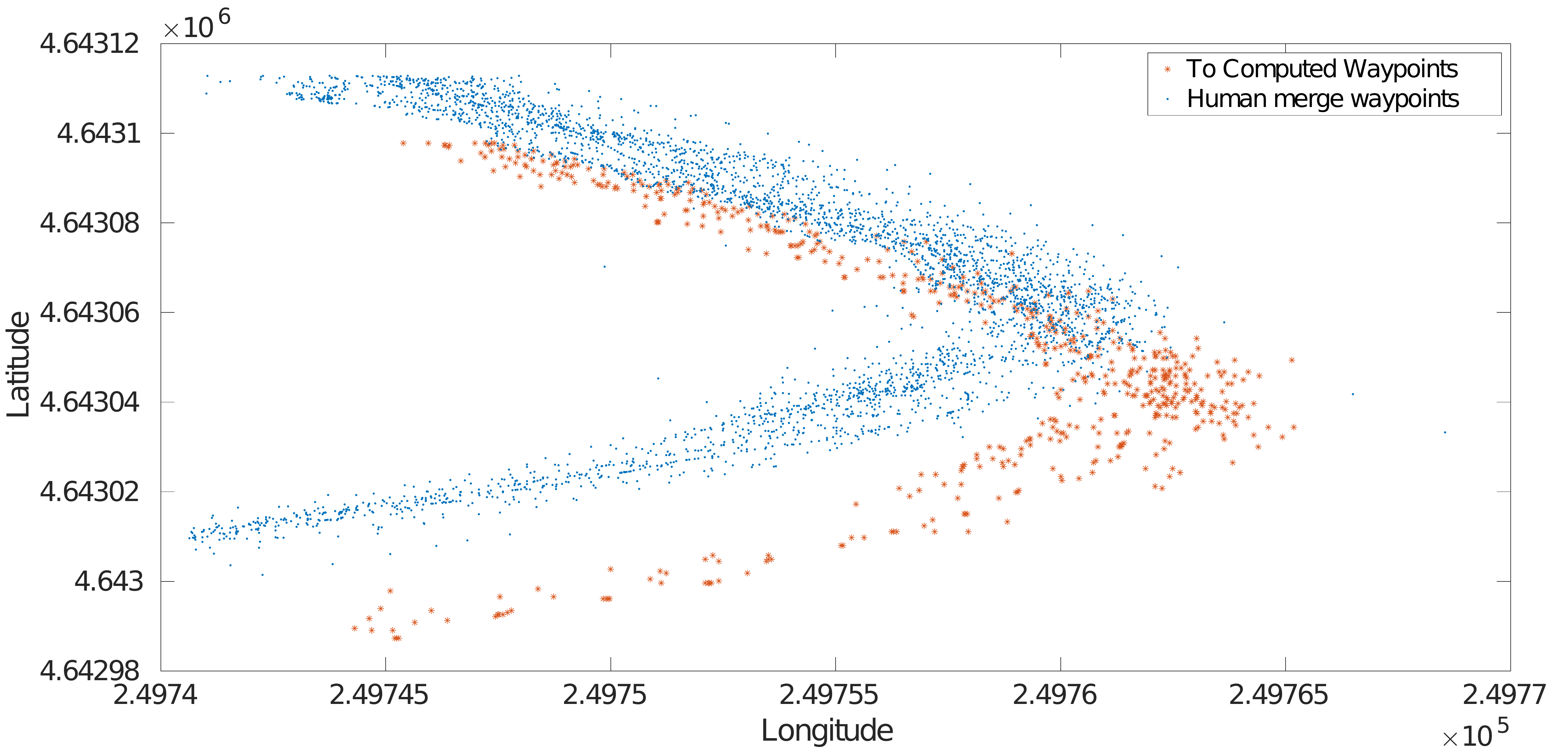}}
    \caption{Real trajectories from the test track.}
    \label{fig:real_scenario}
\end{figure}

The test track for the cooperative driving scenario is shown in Fig. \ref{fig:real_scenario_b}. For the tests, three connected vehicles are used (merging, following and preceding) while the fourth is unconnected. The lane merge coordination elements presented in Fig. \ref{fig:LaneMergeCoordination} were implemented using the \textit{Image Recognition} system from \cite{kai}, the \textit{GDM}, the \textit{\gls{v2x}}, the \textit{\gls{5g} Network} and the edge cloud from \cite{v2x_systems}. The \textit{Traffic Orchestrator} (Fig. \ref{fig:proposed_architecture}) was implemented using the \gls{dueling-dqn} model with positive reward, since it is more reliable in terms of predicting acceleration and heading, compared with the \gls{ml} classifiers. The evaluation consists of two main results: the human likeness of the predicted trajectory and the predicted acceleration.

In order to compare live \textit{Traffic Orchestrator's} predicted trajectories and human trajectories, a preliminary test was implemented without the interaction of the \textit{Traffic Orchestrator}, while the rest of the architecture was up and running. To do that, several merges were performed in the test track to get statistics of those merges on the road. During those tests, the \textit{Image Recognition} and connected vehicles were sending data to be stored by the \textit{Data Fusion}. These preliminary tests are the human merges that are used to compare the predicted trajectory and the human likeness of the manoeuvre. Fig. $\ref{fig:human_vs_to_real}$ shows the latitude and longitude of the merging scenario for human and predicted way points. The general shape of the merge has been detected and successfully predicted by the \textit{Traffic Orchestrator} corresponding to the road architecture (Fig. \ref{fig:real_scenario}), this is an indicator that the model can adapt and generalise to unseen real world scenarios, on the other hand, it is clear that there is a bias from the predicted way points. During the tests, the \gls{rl} model required a high synchronisation level with other components in real-time, therefore, high frequency, low latency and great precision were required to ensure that the \textit{Traffic Orchestrator} could feed the correct data to the \gls{rl}. As such, the bias could stem from the minor delays the architecture incurred.

\begin{figure}[!t]
    \centering
    \includegraphics[width=0.5\textwidth]{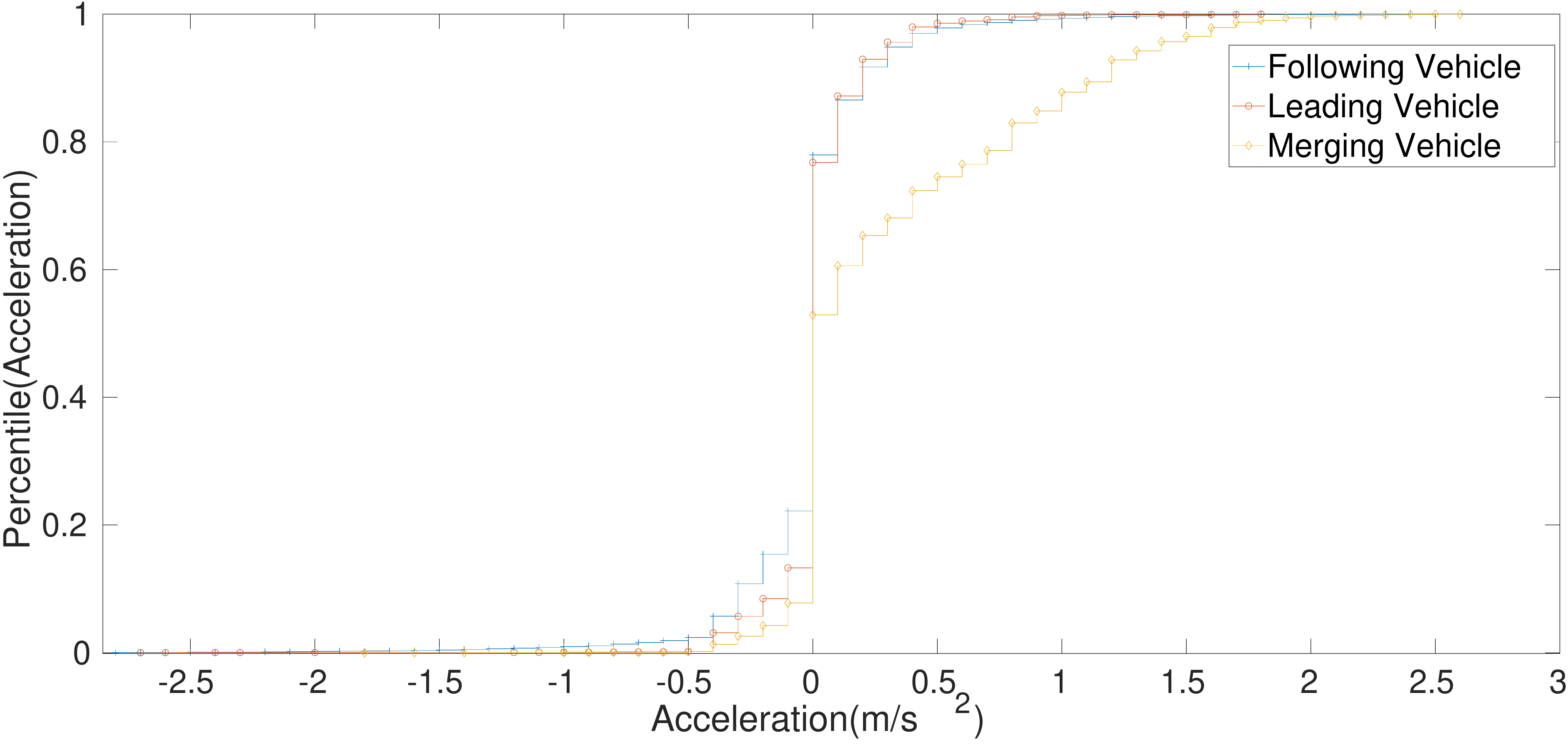}
    \caption{\gls{ecdf} of acceleration during merging.}
    \label{fig:acceleration_real}
\end{figure}

On the other hand, Fig.$\ref{fig:acceleration_real}$ presents the \gls{ecdf} of the acceleration for each vehicle on the test track. The majority of the acceleration values lied in the range $0-2 \; m^2/s$ which is a good value for acceleration that mimics a human driver during a lane merge. This is a good indicator that the \textit{\gls{to}} predicted acceleration values that provided a smooth trajectory recommendation during the manoeuvre. From the following vehicle's point of view, the recommendations given had the intended purpose of slowing down the following vehicle to create a larger gap in between the vehicles on the target lane, for a safer and smoother merge experience.

\subsection{Lesson learnt}

In general, lane merge is a challenging use case in cooperative driving, since there are multiple connected and unconnected entities involved and should be coordinated. As a result, there is also a large amount of architecture components to interact with each other. To this end, the first lesson learnt is to avoid dependency on external components, where possible. The \textit{Traffic Orchestrator} faced a few challenges during the testing phase, due to its strong dependency on data quality, synchronisation and availability of other components which could not be avoided.

A granular design for developing \gls{ml} applications can improve performance, for instance, creating independent and custom microservices for predicting each desired feature such as merges, acceleration and heading. This could require the addition of a management layer on top of them. By doing so, hyperparameters can be adjusted according to the needs of each model. This will also speed up debugging process and future improvements. Even though model's performance is very important for the \textit{Traffic Orchestrator}, in real evaluation tests, defining automotive \glspl{kpi} is a key step to measure general performance across the architecture. In the lane merge scenario, the inter-vehicle distance, manoeuvre length and acceleration are suggested. 

In terms of network performance, the round-trip-time of a trajectory recommendation is a good indicator but it requires a high level of synchronicity across every component, in order to be measured. In this context, default \gls{tcp} configuration of certain mechanisms like delayed acknowledgements and the congestion control algorithm need to be adjusted for this particular scenario, so that the optimal network performance can be achieved. On the other hand, implementing a performance monitoring system is very helpful for noticing, understanding and locating performance issues. Moreover, it provides a new rich source of real data.

%% file: text/conclusion.tex
In this paper, we presented a cooperative driving scenario, i.e., lane merge coordination. The model is based on central decision making entity at the edge cloud, communicating with the vehicles using a \gls{5g} network. Different machine learning techniques, including Random Forest, K-Nearest Neighbours, Decision Tree, Gradient Boosting and \gls{dueling-dqn}, are used to combine information from connected vehicles as well as roadside cameras, and predict safe trajectories in timely manner. Our results show that the \gls{dueling-dqn} model perform best by providing more human-like trajectories. Predicted trajectories provide smooth driving experience of acceleration in the range of $0-2 \; m^2/s$. The \textit{Traffic Orchestrator} achieved a real-time processing for generating safe and successful lane merges, however future works need to be carried out in order to improve \textit{Data Fusion's} processing time and to enhance transport protocol performance. 

%% file: text/acks.tex
This work has been performed in the framework of the H2020 project 5GCAR co-funded by the EU. The views expressed are those of the authors and do not necessarily represent the project. The consortium is not liable for any use that may be made of any of the information contained therein. This work is also partially funded by the EPSRC INITIATE EP/P003974/1 and The UK Programmable Fixed and Mobile Internet Infrastructure. 